\pdfoutput=1

\documentclass[11pt]{article}

\usepackage[]{EMNLP2023}

\usepackage{times}
\usepackage{latexsym}
\usepackage{comment}

\usepackage[T1]{fontenc}

\usepackage[utf8]{inputenc}

\usepackage{microtype}
\usepackage{algorithm}
\usepackage{algpseudocode}
\usepackage{eqparbox}
\usepackage{microtype}
\usepackage{arydshln}
\usepackage{enumerate}
\usepackage{enumitem}
\usepackage{graphicx}
\usepackage{booktabs}
\usepackage{amsmath}
\usepackage{dblfloatfix}
\usepackage{tabularx}
\usepackage{cleveref}
\usepackage{natbib}

\usepackage{inconsolata}

\title{Explicit Diversity Conditions for Effective Question Answer Generation with Large Language Models}

 \author{Vikas Yadav$^\dagger$, Hyuk Joon Kwon$^\ddagger$, Vijay Srinivasan$^\ddagger$, Hongxia Jin$^\ddagger$ \\
  ServiceNow$^\dagger$, Samsung Research America$^\ddagger$, USA \\
  {\tt vikas.yadav@servicenow.com, bluecube246@gmail.com}, \\ \ {v.srinivasan,hongxia.jin\}@samsung.com}}

\begin{document}
\maketitle
\begin{abstract}

 Question Answer Generation (QAG) is an effective data augmentation technique to improve the accuracy of question answering systems, especially in low-resource domains. While recent pretrained and large language model-based QAG methods have made substantial progress, they face the critical issue of redundant QA pair generation, affecting downstream QA systems. Implicit diversity techniques such as sampling and diverse beam search are proven effective solutions but often yield smaller diversity. We present explicit diversity conditions for QAG, focusing on spatial aspects, question types, and entities, substantially increasing diversity in QA generation. Our work emphasizes the need of explicit diversity conditions for generating diverse question-answer synthetic data by showing significant improvements in downstream QA task over existing widely adopted implicit diversity techniques.
 In particular, generated QA pairs from explicit diversity conditions when used to train the downstream QA model results in an average 4.1\% exact match and 4.5\% F1 improvement over QAG from implicit sampling techniques on SQuAD\textsubscript{DU}. 
 Our work emphasizes the need for explicit diversity conditions even more in low-resource datasets (SubjQA), where average downstream QA performance improvements are around 12\% EM.

\end{abstract}

\section{Introduction}

\begin{figure*}
\includegraphics [width=2.05\columnwidth, height=265pt]{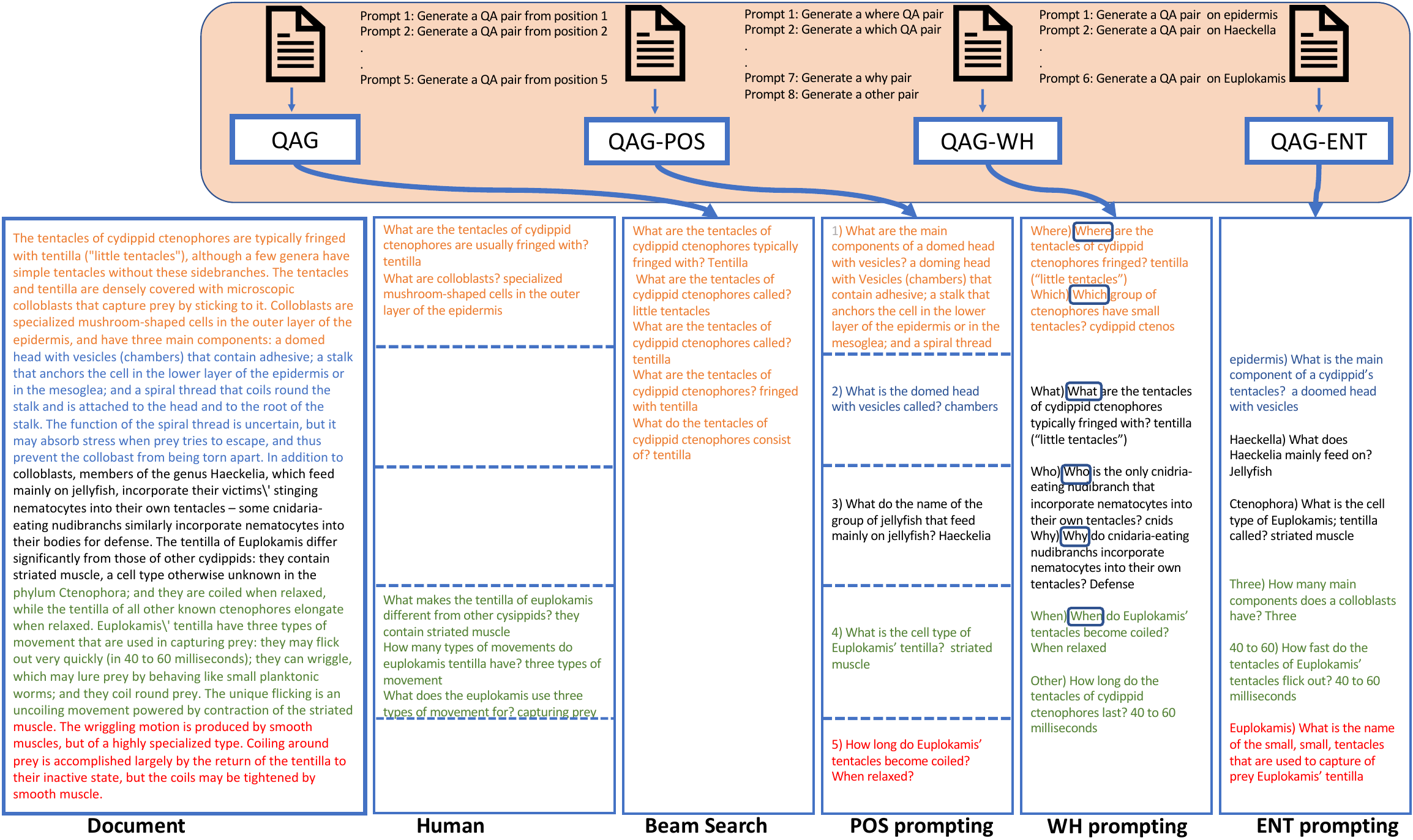} \\[\abovecaptionskip]
\vspace{-6mm}
 \caption{\footnotesize A sample input passage and QA pairs generated by human annotators, nucleus sampling based beam search and our explicit diversity prompting techniques. Different colors in the document text depict the 5 different positions. QA pairs from specific positions are depicted in the same font color and \textit{WH} question types are indicated in blue bounding boxes. Example of each explicit diversity prompts are shown in the top block.}
  \label{fig:examplefig}
  \vspace{-2mm}
\end{figure*}

Annotating QA pairs is costly, tedious, and constrained to annotators' limited coverage of the input document which often leads to lower QA performance in low resource domains \cite{rajpurkar-etal-2016-squad, bartolo-etal-2020-beat, yadav2019quick}. Recent QAG methods, particularly neural pretrained language models (PLM) and large language models (LLM), have generated high-quality synthetic QA pairs leading to strong downstream QA performance \citep{du-cardie-2018-harvesting, puri-etal-2020-training, stasaski-etal-2021-automatically}. 
It is reported that even these prominent neural QAG methods suffer from repeated redundant generation, even after utilizing several implicit techniques for diverse generations such as nucleus, topK sampling, and diverse decoding methods \cite{shao-etal-2017-generating, sultan-etal-2020-importance}. Our work evaluates diversity of such widely adopted implicit techniques for QAG and show the QA generations to be still largely redundant, affecting downstream QA performance. We conjecture that artifacts in human annotations of the training data leads to QAG redundancy. For example, 71\% of the questions in the benchmark QAG dataset SQuAD\textsubscript{DU} are annotated from the first half of the document, and 73\% of the questions are of the type \textit{who, how, what, and why}. As shown in \cref{fig:examplefig}, human annotators have annotated QA pairs only from the top and 4/5th position of the passage and only \textit{what} and \textit{how} questions. Training on such skewed dataset may overfit neural QAG methods on numerous annotator artifacts, thus reducing diversification effectiveness of implicit sampling techniques. 


 Our work focuses on explicit diversity conditions where we present three types of explicit prompts, conditioning QAG on (1) various positions (\textit{POS}) within the input document from where QA pairs are generated, (2) 8 types of \textit{WH} questions for generating questions of different types, and (3) questions based on different named entities (\textit{ENT}). As shown in upper block of \cref{fig:examplefig}, these explicit diversity conditions are concatenated as prompts to the input document for diverse QA generation. These explicit conditions can also be easily combined to one another for jointly prompting QAG models, especially the LLM based ones where we observed the best downstream QA performance (\cref{evalSection}). 
 Our work primarily focuses on establishing the importance of adding diversity conditions explicitly over the widely adopted implicit sampling techniques. The clear benefits of explicit prompting based QAG are highlighted with improved downstream QA performance (\cref{evalSection}) and coverage of diverse information (\cref{Overlap_analyses}) from the input document.
 Our key contributions are:
\vspace{-2mm}

{\flushleft {\bf (1)}} We study diversity of implicit sampling techniques and compare them with several explicit diversity conditions for QAG. The synthetic QA pairs generated from our explicit diversity conditions significantly improve the downstream QA performance outperforming implicit sampling techniques by 4.1\% EM and 4.5\% F1 on widely studied SQuAD\textsubscript{DU} dataset \cite{du-etal-2017-learning}. The improvements from our explicit conditions drastically exceed in the multi-domain low resource SubjQA dataset \cite{QGbench} with improvements of 12\% F1 score. 
 

\vspace{-2mm}

{\flushleft {\bf (2)}} 
Our explicit diversity prompts show substantial diversity improvements, resulting in only 30\% token overlap among generated QA pairs from the input document, compared to the 64\% overlap in QA pairs from implicit sampling-based QAG. The coverage of information from the input document in terms of position, question type, and named entity attributes is also considerably higher in generated QA pairs from explicit diversity prompting over implicit sampling techniques.

\section{Related Work}




Recent studies have highlighted redundancy in neural-QAG approaches and while some widely adopted diverse sampling and beam decoding methods \cite{sultan-etal-2020-importance, holtzman2019curious, vijayakumar2018diverse} have shown improvements, these implicit techniques only moderately enhance the diversity of generation (see \cref{tab:overlap_analysis}). Furthermore, implicit sampling techniques lack precise control of QAG for accessing specific information from the input document. For example, QAG models using nucleus sampling or diverse decoding would still generate QA pairs from a random position of the document or of random WH question type. In contrast, our explicit prompting techniques offer high control over QA generation, allowing selection from specific positions or named entities in the input document, and the types of questions (shown in last 3 columns of \cref{fig:examplefig}).



Many previous QAG methods can be broadly categorized as either explicit or implicit techniques. For instance, \citet{zhou2019question} is analogous to our explicit WH-type QAG model, while answer selector modules \cite{yao-etal-2022-ais, back-etal-2021-learning, puri2020training}, which select answer spans to condition QG, are analogous to our entity-conditioned QAG method. On the other hand, sampling, beam search and additional embedding \cite{lee-etal-2020-generating} based approaches can be grouped under implicit diversity conditions. From these, we mainly focus on widely adopted sampling and diverse decoding methods as implicit diversity baselines(\cref{samplingtechniques}). Our work primarily focuses on comparing these two broad directions of explicit versus implicit diversity methods by showing their impact on diverse generation, downstream QA performance, and information coverage from the input document.  We show experiments on the standard question generation benchmark - QGbench \cite{QGbench}. QGbench authors highlighted higher performance from pretrained-LMs over RNN based models. For fair comparisons, we implemented both explicit and implicit sampling techniques on the same base PLM (BART \cite{lewis-etal-2020-bart}) and LLM (LLaMa-7B \cite{touvron2023llama}) models. Further, \citet{ushio_empirical} showed end-to-end QAG as the best setting where both question and answer are generated as a single output, in a single generation step. We use the same setting throughout our experiments.

\section{Approach}

The task of QAG is to generate QA pairs given an input document. Formally, given a document D which contains M tokens, $D = (d_{1}, . . . , d_{M})$ and $i^{th}$ QA pair has $t$ number of tokens i.e., $qa^i = (qa^i_{1}, . . . , qa^i_{t})$, the task is formulated as a conditional sequence generation at the token level i.e., $P(qa^i_{k}|qa^i_{1}...qa^i_{k-1},d_{1}, . . . , d_{M})$. We model the conditional probability $P(qa|D)$ using 1) BART, a PLM that achieves best results on QGbench 
and 2) LLaMa-7B, a decoder only LLM. Our three explicit diversity conditions are described below.


\subsection{Explicit diversity prompts}
{\flushleft {\bf POS prompting}}:
  We consider 5 splits of the input document based on its total word count. For example, if a document contains 400 tokens, each split will cover 80 tokens each. QAG\textsubscript{POS} is then conditioned on positions of each of the 5 splits, thus encouraging generation of QA pairs from 5 different positions\footnote{We tried different number of positions from the document $\in {2,5,10}$ and found the best QAG with 5 positions.}. In particular, we explicitly prompt the QAG\textsubscript{POS} model to generate $qa$ from $pos$ position of the document where $pos\in{1,2,3,4,5}$.
    
\noindent
\begin{align}
qa_{pos} \sim P(qa|D, pos)
\label{eqn:BARTpos}
\end{align}
 For example, to generate a QA pair from the 2nd position of the document (shown by blue font in the 1st column of \cref{fig:examplefig}), we prompt: "Generate a QA pair from position $2$" to our QAG\textsubscript{pos} model. Splitting of the input document based on word count is a bit rigid segmentation. However, we observed that even with such rough segmentation, QAG\textsubscript{pos} model is able to learn approximate alignment between position of the input document and its corresponding generated QA pair. During training, we use the start offset of the human annotated answer to determine which position split the QA pair was annotated from. During inference, we generate 5 QA pairs from all the 5 different positions of the document. 


  {\flushleft {\bf WH prompting:}} Similar to POS prompting, we condition on the $wh$ question type 
    where $wh \in$ \{\textit{"where", "which", "when", "what", "who", "how", "why"} \} to encourage the QAG\textsubscript{WH} model to generate different types of questions.  
    \begin{align}
qa_{wh} \sim P(qa|D, wh)
\label{eqn:BARTwh}
\end{align}

During training of QAG\textsubscript{WH}, we use the $wh$ type from human annotated questions and during inference, we simply generate QA pairs by conditioning on all 7 $wh$ types. If the annotator's question did not have any $wh$ type, then we consider it as \textit{"other"}. As shown in 2nd last column of \cref{fig:examplefig}, prompting QAG\textsubscript{WH} with $wh$ generates diverse QA pairs with different question type.

  {\flushleft {\bf ENT prompting:}} QAG\textsubscript{ENT} is conditioned on named entities in the input document to generate entity-specific QA pairs. During training, we select named entities present in the human annotated QA pairs and the input document, identified using the SpaCy NER tagger with 18 entity classes from OntoNotes~\cite{weischedel2011ontonotes}. 
  During inference, we split the document into individual sentences and select the longest named entity from each sentence to use in the prompt. The named entity-conditioned prompt, along with the input document, generates a QA pair for that specific entity. As shown in \cref{fig:examplefig}, QAG\textsubscript{ENT} generates diverse QA pairs by conditioning on different entities from the input document.

\subsection{Combined prompts}
\label{learnedprompts}
Our three base diversity prompts can be rigid sometimes; for example, a specific WH question may not be feasible for a particular document. To address this issue, we propose a two step process, with the first step being $wh$ question type prediction given a position or entity. For example, given $pos=2$, a trained $wh$ predictor model predicts "what" type of question in the 1st step. In the 2nd step, QAG model generates a "what" type question from the 2nd position of the document. This two step process of combining $wh$ type with position and entity diversity conditions is explained below.

\begin{itemize}[noitemsep,leftmargin=0mm]
    \item \textbf{Position-based question type generator}: We train a separate BART model to generate a list of relevant WH-types (or 'none' if no QA is possible) for a specific position in the input document. Then, we generate QA pairs conditioned on both the specified position and the predicted WH types. 

     \begin{align}
wh \sim P(wh|D, pos) \\
qa_{pos,wh} \sim P(qa|D, pos, wh)
\label{eqn:BARTQAGWHPOS}
\end{align}

    \item \textbf{Entity-based question type generator}: Similarly, we predict potential $wh$ question types for the selected entity from the input document. We then generate QA pairs given the selected entity and the predicted WH types.
    \begin{align}
wh \sim P(wh|D, ent) \\ 
qa_{ent,wh} \sim P(qa|D, ent, wh)
\label{eqn:BARTQAGentPOS}
\end{align}
\end{itemize}

Please note that this 2 step process for combing different explicit diversity conditions is required only for BART-QAG. As LLMs can follow instructions and generate long sequences \cite{wei2022emergent}, a single prompt for combining two explicit conditions - \textit{"Generate N questions of different question type from different positions"} is given to the LLaMa-7B QAG model. This single prompt based combining of explicit conditions for QAG is referred to as $Combined$ in \cref{tab:questionAperformance}.

\subsection{Implicit Sampling and Decoding}
\label{samplingtechniques}
We considered four widely adopted decoding techniques as implicit diversity baselines: nucleus sampling, top\_k sampling, beam search with sampling, and diverse decoding  \cite{sultan-etal-2020-importance, holtzman2019curious, vijayakumar2016diverse, vijayakumar2018diverse}. In addition, we also assessed these sampling techniques in combination with our explicit diversity conditions. Our position and entity prompt conditioned QAG models performed consistently better with diverse decoding, while WH prompting showed higher downstream QA performance with nucleus sampling.

\vspace{-3mm}

\paragraph{Hyperparameters:} We conducted our experiments with the PyTorch implementation of BART and LLaMa from the Hugging Face Transformers library \cite{wolf-etal-2020-transformers}. For training BART-QAG, the final hyperparameters were epoch=4, batch size=16, learning rate=3e-5, and adam optimizer on V100 GPUs. The remaining hyperparameters were used as default suggested in \cite{QGbench}. We used 4 A100 GPUs to finetune our LLaMa-QAG models using following hyperparameters: batch size=4, learning rate=2e-5, epoch=3, and float32=True. For training the BERT-large-uncased-wwm QA model, the final hyperparameters were epoch=2, learning rate=3e-5, seq length=384, batch size=8, and stride=128

\section{Evaluation}
\label{evalSection}
We evaluate the impact of our explicit diversity prompts on downstream QA task on the standard datasets from QGbench (i) SQuAD\textsubscript{DU}\cite{du-etal-2017-learning} (ii) and low-resource multi-domain SubjQA \cite{bjerva2020subjqa} which has less than 150 annotations. We trained a BERT-large-uncased-wwm based QA model \cite{devlin-etal-2019-bert} over synthetic QA pairs generated from our explicit conditions and implicit sampling techniques. Each experiments is run 5 times and average QA performance on the SQuAD\textsubscript{DU} test split is reported in \cref{tab:questionAperformance}. We report both the F1 and exact match (EM) using the standard SQuAD evaluation script \cite{rajpurkar-etal-2016-squad}.

\begin{table*}[t]
\centering
\scriptsize
\resizebox{1.6\columnwidth}{!}{
\begin{tabular}{lll|ll|ll}

\toprule
 & &  & \multicolumn{2} {c|}{\textbf{BART}}  & \multicolumn{2} {c}{\textbf{LLaMa-7B}}  \\

\textbf{SRC} & \textbf{I/E} & \textbf{Approach} & \multicolumn{2} {c|}{\textbf{Orig Size}}   & \multicolumn{2} {c}{\textbf{Orig Size}}  \\
    & & & EM & F1 & EM & F1  \\
\toprule

 Syn & I & Greedy &  64.76	& 76.66 & 71.41 & 83.26  \\
 Syn & I & Nucl (0.95) & 64.44 &	77.15 &  71.92 & 83.53 \\
 Syn &  I  & Nucl+TopK & 64.17 & 76.47 & 72.08 & 83.71  \\
 Syn & I & DiverseDec. &  65.21 &	77.37 & 71.87 & 83.66 \\
\midrule
Syn & E & WH Prompt &  67.25 & 79.60 & 72.97 & 84.46\\
 Syn & E & POS Prompt & 69.62	& 81.49 & 72.74 & 84.25 \\
 Syn & E & ENT Prompt & 69.31	& 81.80 & 72.59 & 84.21  \\
\midrule
 Syn &  & POS->WH & 71.77 & 83.30 & - &  - \\
 Syn &  EL & ENT->WH & 70.46 &	81.78 &  - & - \\
  Syn &   & Combined & - &	- &  {\bf73.29} & {\bf84.76}\\
 \midrule
 H+Syn &   & SQ\textsubscript{dev}+WH & 74.30	& 85.62 & 75.11 & 86.42   \\
 H+Syn & & SQ\textsubscript{dev}+POS & 74.53 & 85.61 & {\bf75.76} & {\bf87.15}  \\
 H+Syn & & SQ\textsubscript{dev}+ENT &  73.17 & 85.01  & 75.59 & 87.06  \\
\midrule
 H &  & SQ\textsubscript{dev} & \multicolumn{2} {c|}{EM=74.08} & \multicolumn{2} {c}{F1=85.19}  \\
\bottomrule
\end{tabular}
}
\caption{\label{tab:questionAperformance} \footnotesize Downstream QA performance on the QG-bench SQuAD DU test dataset. We use top\textsubscript{p}=0.95 and topK=30.The third-row block settings refer to the learned combination of diversity conditions (\cref{learnedprompts}) where the first prompt predicts the second potential diversity prompt (separated by ->). I, E, and EL in the 2nd column stand for implicit, explicit, and learned explicit conditons respectively. SQ refers to SQuAD\textsubscript{DU} dev split. Nucl and DiverseDec are short form for Nucleus and DiverseDecoding. The Orig Size indicates that the synthetic data size matches the original training size of SQuAD DU dataset of 10570 QA pairs. The eval dataset for all the rows is the SQuAD DU test split which contains 11877 QA pairs. H and Syn refers to human annotated and synthetic QA dataset.
}
\end{table*}







\begin{table*}[t]  
\centering  
\begin{tabular}{ll|c|llll}  
\toprule  
\textbf{Data} & Eval & \textbf{Human} & \multicolumn{4} {c}{\textbf{BART-QAG}}  \\   
&  & & NS & POS & WH & ENT \\  
\hline  
Books & EM & 6.3 & 7.9 & 14.7 & 11.6 & 20.0  \\  
T,E=92,191 & F1 & 20.3 & 25.9 & 29.4 & 30.1 & 37.9 \\  
\hdashline  
Electronics & EM & 15.2 & 16.4 & 23.6 & 27.9 & 25.7 \\  
T,E=99,238 & F1 & 34.3 & 33.6 & 44.4 & 47.5 & 47.5 \\    
\hdashline  
Grocery & EM & 14.6 & 16.5 & 16.2 & 0.0 & 15.4  \\    
T,E=101,379 & F1 & 31.9 & 32.0 & 31.2 & 16.1 & 31.1  \\  
\hdashline  
Movies & EM & 13.6 & 15.6 & 23.4 & 27.9 & 25.3  \\     
T,E=101,154 & F1 & 30.2 & 30.2 &36.9 & 41.3 & 39.3  \\  
\hdashline  
Restaurant & EM & 8.2 & 0.0 & 6.7 & 12.7 & 26.1 \\  
T,E=129,136 & F1 & 23.9 & 7.1 & 20.3 & 25.9 & 40.3  \\  
\bottomrule  
\end{tabular}  
\caption{\label{tab:LowResourceResults} Downstream QA performance on the SubjQA test dataset of QG-bench. NS refers to nucleus sampling.  Numbers in bold represent the best performance in each column. T and E under each domain refer to the number of QA pairs in training and evaluation split respectively. Please note that BART-QAG model generates the same number of QA pairs as annotated in Human (Hum) sets in each domain.}  
\end{table*}

\begin{enumerate}

\item {\flushleft {\bf Downstream QA performance}}: QA model trained on data from explicit diversity prompted QAG models (row-block 2 of \cref{tab:questionAperformance}) achieve, on average, 4.1\% higher EM and 4.5\% higher F1 scores compared to the implicit sampling techniques (row-block 1 of \cref{tab:questionAperformance}) in our BART-QAG methods. These empirical evidences highlight the importance of explicit diversity conditioning for effective QAG. 
Performances improve further (row-block 3) when diversity conditions are combined in a learned setting (\cref{learnedprompts}), suggesting the need for a learned module to capture complex relationships between the three explicit diversity conditions. Interestingly, combining multiple explicit conditions in a single prompt for LLaMa-QAG (denoted by \textit{Combined} in \cref{tab:questionAperformance}) results in the best downstream QA performance.


\item{\flushleft {\bf BART vs LLaMa}} - QA pairs generated from LLaMa-QAG consistently lead to better downstream performance than BART-QAG. As expected, the improvements are smaller from explicit conditions in LLaMa-QAG because of their extensive pretraining leading to more qualitative generations. LLaMa-QAG synthetic QA data from explicit conditions almost matches performance from human annotated SQuAD\textsubscript{DU} dataset (within 0.4\% F1). Interestingly, just appending QA pairs from explicit conditioned LLaMa-QAG to human annotated SQuAD\textsubscript{DU} leads to ~2\% F1 improvement (row block 4), resulting in best performance of 87.2\% F1 in downstream QA task. This highlights the benefits of combining high-quality diverse synthetic data to existing human annotated QA pairs. Although average improvements in downstream QA tasks from explicit diversity prompts are smaller in LLaMa-QAG, the generated QA pairs still have higher coverage and diversity compared to implicit sampling techniques (discussed in \cref{Overlap_analyses}). 
\item{\flushleft {\bf Low resource QAG}} - In \cref{tab:LowResourceResults}, we observed substantially higher performance improvements with our explicit diversity-conditioned BART-QAG on the SubjQA datasets. Particularly, synthetic data from explicit diversity-conditioned BART-QAG resulted in a 7\% EM and 10\% F1 improvement over implicit nucleus sampling based QA data. Interestingly, explicit-conditioned QA pairs lead to on-par or higher performance when compared to small-sized human annotated data of SubjQA. Thus, emphasizing the importance of explicit diversity conditions even more in low-resourced domains. 
\end{enumerate}

\section{Overlap and Coverage Analyses}
\label{Overlap_analyses}

\begin{table}
    \scriptsize
    \resizebox{\columnwidth}{!}{
   \centering
      \begin{tabular}{@{} c|c|ccc|c}
      \toprule
   Analysis & \textbf{Overlap} & \multicolumn{3} {c}{\textbf{Coverage}} & \textbf{Time} \\
   & & POS & WH & ENT & (ms)  \\
    \hline
Greedy & 63.07 & 36.84 & 31.33 & 32.18 & 223.1\\
Nucl (0.95) & 57.44 & 57.15 & 45.59 & 29.80 & 372.1 \\
Nucl+TopK & 59.93 &  58.62 & 48.23 & 30.21 & 451.4\\ 
DiverseDec & 46.85 & 49.83 & 42.76 & 35.38 & 388.2\\
\hdashline
POS Prompt & 36.10 & {\bf 77.56} & 34.62 & 50.62 & 231.5 \\
WH Prompt & {\bf 30.67} & 60.41 & {\bf 97.81}  & 48.06 & 218.7 \\
ENT Prompt & 34.59 & 75.89 & 55.34 & {\bf 63.90} & 227.9\\
\hdashline
Human & {\bf 28.04} & 65.82 & 56.32 & 44.96 & - \\ 
    \end{tabular}
    }
    \caption{ Pairwise lexical overlap between generated QA tokens, their coverage, and average generation time for 5 QA pairs from SQuAD\textsubscript{DU}. } 
    \label{tab:overlap_analysis}
\end{table}



We compute the lexical token overlap between the generated QA pairs for each document. For this analysis, we generated 5 questions with each approach and report the average pairwise token lexical overlap between all $\binom{5}{2}$ QA pairs over SQuAD\textsubscript{DU} dev split. As shown in \cref{tab:overlap_analysis}, there is a substantially higher average token overlap of 63.1\% between QA pairs generated by greedy beam search clearly highlighting the diversity problem. Nucleus sampling and diverse decoding have comparatively lower overlap ratios (57.4 and 49.8) but are still substantially higher than our techniques suggesting the need of explicit diversity conditioning. $WH$ explicit prompting results in the lowest average token overlap of just 30.7 indicating its effectiveness in diverse QAG. It is worth noting that the overlap of human annotated QA pairs were low because there were $\le2$ QA pairs annotated for majority of the input documents. 

We also compute the average lexical coverage of the 5 generated QA pairs by assessing answer text position (POS), entity and \textit{wh} in question text (denoted by ENT and WH in \cref{Overlap_analyses}). For example, we compute answer position coverage of the generated QA pairs from position 1 to 5. If generated QA pairs have answers only in 4 of the 5 splits, POS coverage will be 80\%. QA pairs generated by our explicit diversity-conditioned methods have substantially higher coverage compared to all the implicit sampling baselines. Unsurprisingly, BART\textsubscript{POS}, BART\textsubscript{WH}, and BART\textsubscript{ENT} have the highest average lexical coverage of spatial, \textit{wh} question type, and named entity in the input document respectively. We also calculate the average generation time of 5 QA pairs per input document (last column of \Cref{tab:overlap_analysis}) highlighting explicit diversity prompts are also much faster than selecting multiple beams and other diverse decoding techniques. 

\vspace{-2mm}

\paragraph{Conclusion:} We presented a detailed study of implicit versus explicit conditioning techniques for diverse QA generation, highlighting lack of diversity in generations from implicit techniques. Our work empirically shows the clear benefits of explicit diversity conditions with substantial improvements in diverse generations, downstream QA task, and information coverage from the input document. We also show that the concatenation of explicit conditioned based diverse synthetic QA pairs to human annotated datasets leads to further improvement in downstream QA performance. Overall, our presented findings suggest the need of utilizing more of explicit diversity conditions over the existing popular diversity sampling techniques, especially in low resource settings.

\section{Future Work}

We focus on the standard and more mainstream QAG task from QGBench but our proposed techniques can be easily extended to other complex QA tasks such as multi-hop QA \cite{yadav2020unsupervised}. Similarly, our explicit diversity techniques can be extended to other text generation tasks such as conversational QA, dialogue, answer summarization etc \cite{reddy-etal-2019-coqa, wu-etal-2022-qaconv}.

In case of position diversity of generated QA pairs, input documents can be longer (or shorter). Although we had tried splits $\in$ \{2,5,10\}, the number of position splits can be variably selected depending on its length in future works. 
In \Cref{Overlap_analyses}, we studied diversity in terms of overlap and coverage of information via simple lexical matching. As future work, the embedding representations of the generated questions throughout QAG model layers can also be evaluated to further understand the effects of training with explicit diversity conditions \cite{yadav2021if}. 



\section{Ethical consideration }

We simply utilize benchmark datasets from QGbench and existing PLM and LLM models like BART and LLaMa-7B. Our presented methodologies and comparitive studies do not induce any biased or harmful content. We believe, similar to the other LLM and PLM based systems, risks depends on the underlying LLM from its pretraining. A careful selection of input documents for QAG and unbiased LLMs or PLMs would ensure safe QA generations from either explicit or implicit techniques. To the best of our knowledge and review of nearly 200 generated QA pairs, we did not find any harmful or biased QA generations.

\bibliography{custom}
\bibliographystyle{emnlp2023}

\appendix

\end{document}